\documentclass[10pt,twocolumn,letterpaper]{article}

\usepackage{iccv}
\usepackage{times}
\usepackage{epsfig}
\usepackage{graphicx}
\usepackage{amsmath}
\usepackage{amssymb}
\usepackage{multirow}
\usepackage{adjustbox}
\usepackage{graphicx}
\usepackage{subcaption}
\usepackage{mathtools}
\usepackage{bm}
\usepackage[accsupp]{axessibility}

\usepackage[pagebackref=true,breaklinks=true,letterpaper=true,colorlinks,bookmarks=false]{hyperref}

\iccvfinalcopy 

\def\iccvPaperID{05} 

\ificcvfinal\fi

\begin{document}

\title{SlowFast Rolling-Unrolling LSTMs for Action Anticipation in Egocentric Videos}

\author{Nada Osman, Guglielmo Camporese, Pasquale Coscia, Lamberto Ballan
\\
Department of Mathematics ``Tullio Levi-Civita''\\
University of Padova, Italy\\
{\tt\small \{nadasalahmahmoud.osman, guglielmo.camporese\}@phd.unipd.it}\\
{\tt\small \{pasquale.coscia, lamberto.ballan\}@unipd.it}\\
}

\maketitle
\ificcvfinal\thispagestyle{empty}\fi

\begin{abstract}
Action anticipation in egocentric videos is a difficult task due to the inherently multi-modal nature of human actions. Additionally, some actions happen faster or slower than others depending on the actor or surrounding context which could vary each time and lead to different predictions. Based on this idea, we build upon RULSTM architecture, which is specifically designed for anticipating human actions, and propose a novel attention-based technique to evaluate, simultaneously, slow and fast features extracted from three different modalities, namely RGB, optical flow and extracted objects. Two branches process information at different time scales, i.e., frame-rates, and several fusion schemes are considered to improve prediction accuracy. We perform extensive experiments on EpicKitchens-55 and EGTEA Gaze+ datasets, and demonstrate that our technique systematically improves the results of RULSTM architecture for Top-5 accuracy metric at different anticipation times.
\end{abstract}

\section{Introduction}
Human action anticipation~\cite{9084270, furnari2018Leveraging, Vondrick} is a popular research topic in computer vision due to a wide range of involved applications. For example, assistive robotic platforms~\cite{Hema, DBLP:conf/icra/SchydloRJS18} need to anticipate human movements to correctly perform their tasks when multiple people are present in the same environment. Similarly, advanced video-surveillance systems~\cite{liang2019peeking} require to anticipate human motion to promptly provide timely assistance. In this context, egocentric videos have provided a considerable amount of information to be used for training action anticipation models thanks to low-cost wearable devices which offer different streams to be used~\cite{munro20multi, Song}, \eg, RGB videos, audio or depth data. 

State-of-the-art approaches~\cite{tab, Wang_Wu_Zhu_Yang_2020} are mainly based on attention mechanisms to efficiently extract relationships across subsequent frames at a specific frame rate. Nevertheless, action speed may differ based on the actor, surrounding environment and action itself. 

To anticipate future actions, two main factors should be taken into account: window size (\ie, number of current and past actions to be considered) and processing frame rate (\ie, quantity of information to be extracted from each action). While the former is typically fixed for a fair results comparison, the latter can be arbitrary selected. In this case, a different choice of this parameter may lead to completely different results. We demonstrate that, if multiple streams of the same modality is provided to an action anticipation model, it is able to appropriately select which stream to focus on and improve its predictive capabilities leading to a better generalization.\\

\begin{figure}[t!]
    \centering
    \includegraphics[width=\columnwidth]{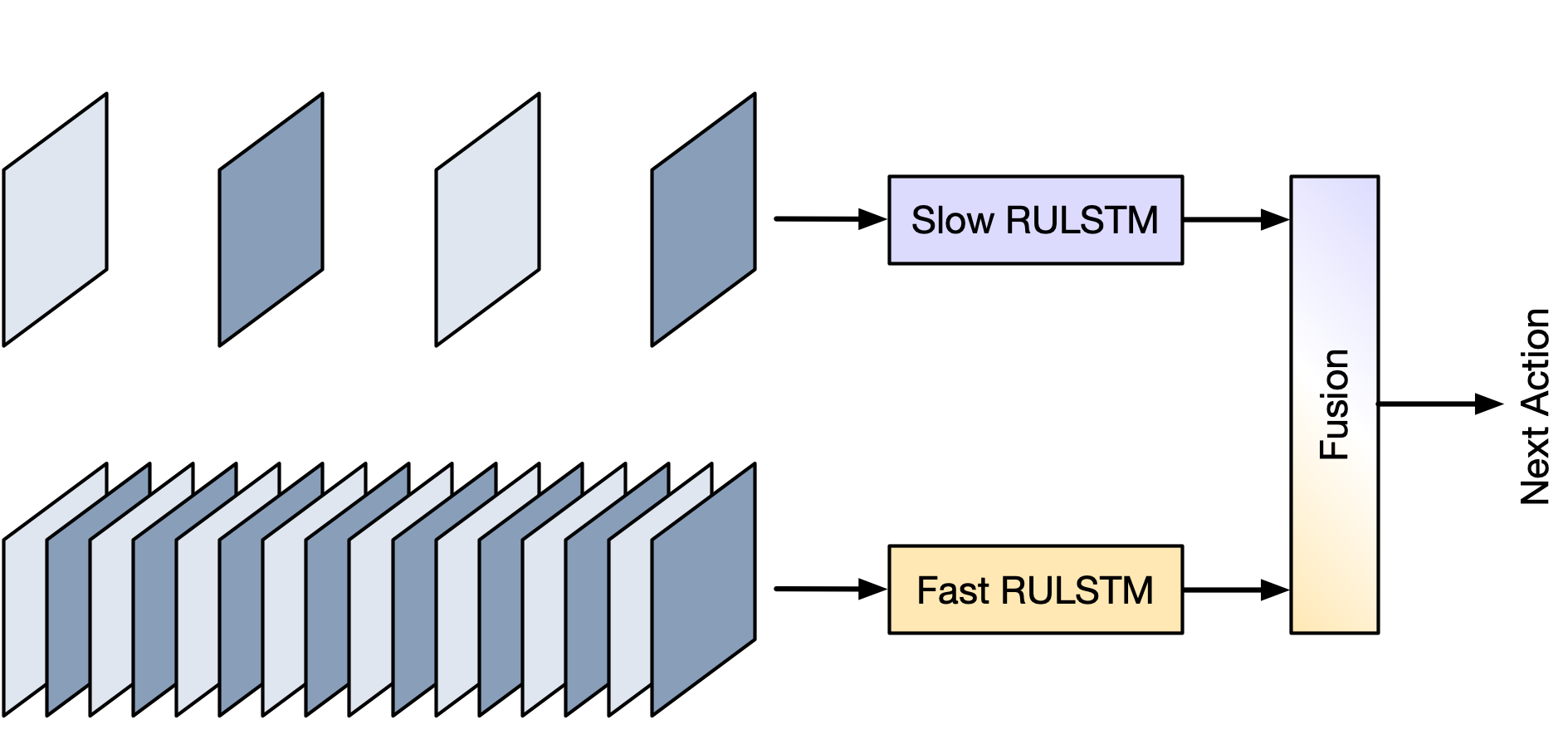}
    \caption{Human actions happen at different speeds requiring a multi-scale approach for better predicting future behaviors. We propose a Slow-Fast RU-LSTM model containing two branches, namely \textit{slow} and \textit{fast} branch, which learn independently from input videos features at different time scales.}
    \label{fig:slow-fast-model}
\end{figure}

Based on this idea, we propose to consider multiple branches for each input modality which process the corresponding stream at different frame rates. We focus on two popular egocentric datasets, namely EPIC-Kitchens-55 and EGTEA GAZE+. Based on RU-LSTM~\cite{rulstm} model, we propose a slow-fast architecture that learns from input videos at two different scales, as shown in Figure \ref{fig:slow-fast-model}. A slow branch processes input videos with a small frame rate while another branch uses a higher frame rate. In this way, redundant information is discarded for actions that evolve slowly while retained for faster actions. In order to efficiently combine these two branches, we use an attentive-based mechanism which efficiently weights their output scores and provides only one result which is subsequently decoded to extract future actions. We show that our model systematically outperforms state-of-the-art models at different anticipation times.\\

The main contributions of our work can be summarized as follows:
\begin{enumerate}
    \item We propose a multi-scale learning technique that benefits from a slow and fast branch to augment performance of RU-LSTM model;
    \item We perform extensive ablation experiments in order to select the most appropriate frame rates and window sizes;
    \item We conduct multiple evaluation experiments on popular action anticipation benchmarks and also compare different model architectures and slow-fast fusion mechanisms.
\end{enumerate}

\section{Related Work}

\subsection{Action Recognition}
Action recognition consists of predicting a labelled action category assigned to an input video. Learning from videos requires capturing both spatial and temporal information, and several approaches have been proposed to solve this task. A simple modelling strategy is based on extracting spatial features from observed video frames with a 2D Convolutional Neural Network (CNN) and their aggregation at temporal level~\cite{large-scale-video-clf, beyond-short-snippets}, or with Long-Short Term Memory (LSTM) networks~\cite{lstm-cnn-visual-rec, beyond-short-snippets}. Another popular approach exploits 3D CNNs where spatio-temporal information is gradually fused, leading to a better video representation and more accurate results~\cite{quo-vadis, R-2plus1-D, conv3d-1, conv3d-2}. Another successful idea uses two-stream networks where RGB frames and optical flow features are processed providing a more detailed motion information contained in input videos~\cite{2stream, 2stream-fusion, quo-vadis, beyond-short-snippets}. 

Recognizing an observed action is the first step for solving more complex tasks, such as early action recognition, where a future action is predicted using only a partial observation of the input video, and action anticipation, where an action category is predicted using only past observed frames.

\subsection{Action Anticipation}
Action anticipation requires to predict future actions relying only on past video frames~\cite{RED}. Previous works proposed different models for activity anticipation in third-person videos~\cite{when-will-you-do-that, what-will-happen-next, RED, a-hierarchical-repr-future, join-prediction, visual-forecasting} and first-person videos~\cite{forecasting-hand-objects, next-active-objects, inverse-reinforcement, deep-future-gaze, label-smoothing-antic}. In our work, we adopt the formulation presented in~\cite{rulstm}, where an action to be predicted is computed at fixed anticipation times before it starts. This is a challenging task since it involves learning both spatial and temporal relationships among past and future frames. To this end,~\cite{rulstm} proposes an encoder-decoder LSTM-based architecture where past information is firstly summarized, and future actions are then computed leveraging features extracted from past information.

\subsection{Multi-Scale Modelling in Vision}
Multi-scale modelling is a powerful design paradigm that empowers a hidden input representation to be more robust to scale changes with respect to a single-scale modelling approach. This technique can be adopted in both spatial~\cite{multiscale-deblurring, multiscale-largescale} and temporal~\cite{slowfast, tab} domains. Slow-Fast networks~\cite{slowfast} for video recognition builds upon this idea and show to benefit from processing video sequences at slow and fast frame rates with two separate branches that capture patterns at different time resolutions. In our work, we take advantage of this idea aiming at capturing slow and fast features for anticipating future actions.

\section{Proposed Method}

Action anticipation consists of predicting future actions using only visual information extracted from current and past frames. As proposed in \cite{rulstm}, a future action is predicted at the anticipation time of $1~s$ before it occurs, and only video information before this anticipation time can be used for its prediction. 
More specifically, the evaluation protocol requires to anticipate the future action at subsequent time steps in order to evaluate the performance of the model approaching the target action. 

In the following, we briefly summarize our baseline, which constitutes the backbone used at different time granularities, and then present our slow-fast fusion technique. Finally, we describe how our slow-fast approach can be both used for one input modality and multiple modalities using modality attention~\cite{rulstm}.

\subsection{Rolling-Unrolling LSTM}
Our technique is built upon RU-LSTM~\cite{rulstm} model, which processes sequences of feature vectors computed from input video frames. This model defines an encoding stage of $S_{enc}$ steps and an anticipation stage of $S_{ant}$ steps for a total of $\alpha \cdot (S_{enc} + S_{ant})$ seconds, where $\alpha$ is the time interval between two subsequent frames. This model is based on an LSTM-based encoder, named \textit{rolling} LSTM (R-LSTM), and an LSTM-based decoder, named \textit{unrolling} LSTM (U-LSTM). The former summarizes, during the encoding and anticipation stages, past information extracted from input videos and provides to the latter a useful context for predicting the future action. The decoder, in the anticipation stage, receives the representation from the encoder and, using the last observation, computes a plausible distribution over future action classes. The encoding-decoding process is performed for each time step in the anticipation stage, and the network is trained for predicting the actual action label using a cross-entropy loss. To exploit more context and create a more informative hidden representation, RU-LSTM processes multi-modal features which are combined using a mixture-of-experts-based method named Modality Attention (MATT). Since this model shows remarkable performance on predicting future actions from multi-modal input streams, we extend its predictive capability by explicitly designing a multi-scale fusion mechanism able to capture slow and fast features from observed video sequences.


\begin{figure*}[t!]
    \centering
    \includegraphics[width=1.\textwidth]{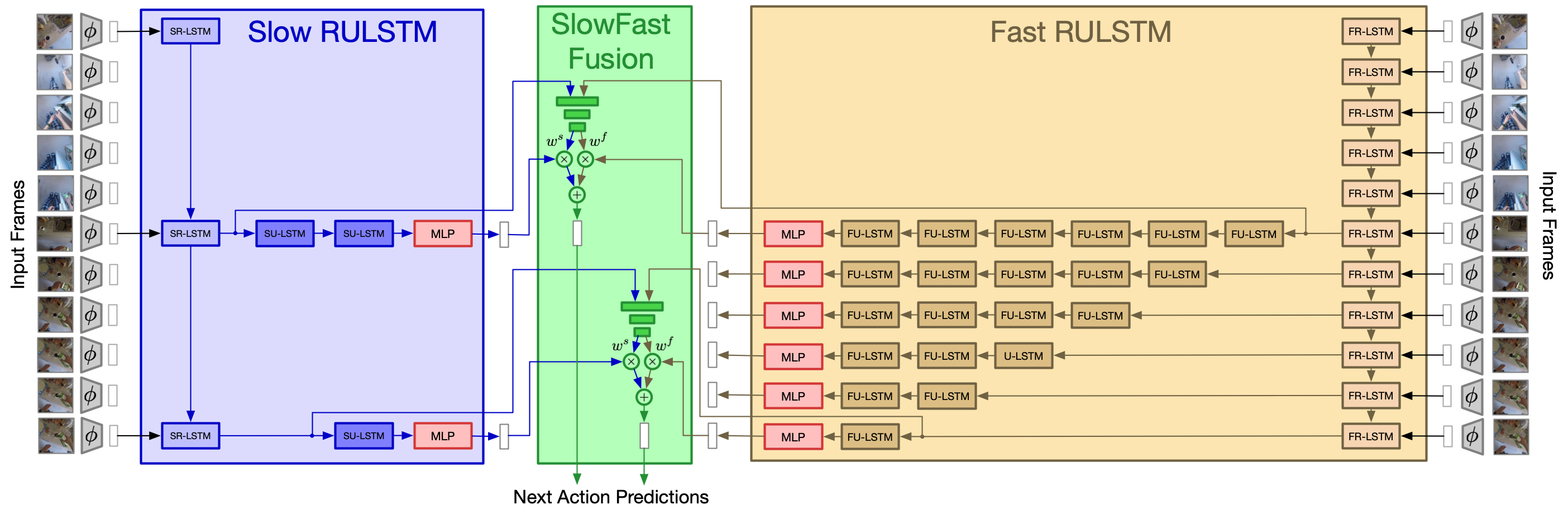}
    \caption{Our SlowFast RULSTM model. Input videos are firstly processed by a CNN feature extractor and then sequence representations are fed to two branches processing information at two different frame rates. Our slow and fast branches are based on RU-LSTM architecture that encodes past information and then decodes future actions. To better capture the correlations in past observed frames, we design a slow-fast fusion mechanism that merges the predictions of these two branches leading to a better accuracy.}
    \label{fig:model}
\end{figure*}

\subsection{SlowFast RULSTM}
\label{sec:slowfast-rulstm}

As depicted in Figure \ref{fig:slow-fast-model}, our SlowFast RULSTM model consists of two branches: a slow branch, that processes input videos using a low frame rate (one frame every $\alpha_s$ seconds), and a fast branch, which uses a high frame rate (one frame every $\alpha_f$ seconds). Our idea is to process input features at different time resolutions in order to capture slow and fast relations between past and future frames. 

Let $\bm{x} \in \mathbb{R}^{T \times C \times H \times W}$ be the input video to be processed and $\bm{z} \in \mathbb{R}^{T \times D}$ the corresponding representation computed at each time step. Given a single input frame $\bm{x}_t$ at time t, $\bm{z}_t = \phi(\bm{x}_t)$ is its related representation where $\phi$ is a CNN feature extractor, and $T = S_{enc} + S_{ant}$ the total sequence length. Our slow branch processes input video frames at $1/\alpha_s$ frame rate while our fast branch at $1/\alpha_f = R/\alpha_s$ with $R = \alpha_s / \alpha_f$ being the ratio between fast and slow frame rates, respectively. Given an internal representation $\bm{z}_t$, the encoder in the fast branch produces feature representations used by the decoder as follows:

\begin{equation}
    \bm{r}^f_t = FR\text{-} LSTM\left( \bm{z}_t, \bm{r}^f_{t-1} \right)
\end{equation}
where $t \in \{1, 2, \dots, T \}$ and $\bm{r}^f_t = (\bm{h}^f_t, \bm{c}^f_t)$ is the state that contains hidden and context vectors of FR-LSTM with $\bm{h}^f_t, \bm{c}^f_t \in \mathbb{R}^{d}$. Our slow branch is similarly defined:

\begin{equation}
    \bm{r}^s_t = SR\text{-}LSTM\left( \bm{z}_t, \bm{r}^s_{t-1} \right)
\end{equation}
where $t = kR + 1$ with $k \in \{0, 1, \dots, \lfloor T/R \rfloor \}$, and  $\bm{r}^s_t = (\bm{h}^s_t, \bm{c}^s_t)$ is the state containing hidden and context vectors of slow R-LSTM with $\bm{h}^s_t, \bm{c}^s_t \in \mathbb{R}^{d}$.
The decoder in the fast branch receives the representations given by the fast encoder and produces the prediction by unrolling the Fast U-LSTM for $T-t+1$ steps as follows:

\begin{align}
    &\bm{u}^f_{t,q} =FU\text{-}LSTM \left( \bm{z}_t, \bm{u}^f_{t,q-1} \right), \\
    &\bm{u}^f_{t,t-1} = \bm{r}^f_t, \ \ \ \ \ \ \ \bm{u}^f_t = \bm{u}^f_{t,T},
\end{align}
where $q \in \{t, \dots, T\}$.
Then, a fast prediction score over all action classes is computed from the output of the decoder with a Multi-Layer Perceptron (MLP) at each time step as $\bm{l}^f_t = MLP(\bm{u}^f_t)$, where hidden and context vectors in $\bm{u}^f_t$ are concatenated.
Similarly, the slow decoder receives the slow encoded features $\bm{r}^s_t$ and produces $\bm{u}^s_t$ by unrolling the Slow U-LSTM and then slow logits $\bm{l}^s_t$ are computed with a MLP. The formulation related to the slow decoding step is as follows:

\begin{align}
    &\bm{u}^s_{t,q} = SU\text{-}LSTM \left( \bm{z}_t, \bm{u}^s_{t,q-1} \right), \\
    &\bm{u}^s_{t,t-1} = \bm{r}^s_t, \ \ \ \ \ \ \ \bm{u}^s_t = \bm{u}^s_{t,T}, \\
    &\bm{l}^s_t = MLP \left( \bm{u}^s_t \right).
\end{align}

After slow and fast logits scores computation, our model fuses the obtained predictions with an attention mechanism. Specifically, given both slow and fast scores ($\bm{l}^s_t$ and $\bm{l}^f_t$), we compute our final merged logits as $\bm{l}_t = w^s_t \cdot \bm{l}^s_t + w^f_t \cdot \bm{l}^f_t$, where $w_s$ and $w_f$ represents slow and fast multipliers that weight slow and fast predictions computed as follows:

\begin{align}
    &\left[ \lambda^s_t, \lambda^f_t \right] = \text{MLP}\left( \left[ \bm{r}^s_t, \bm{r}^f_t \right] \right), \\
    &w^s_t = \frac{e^{\lambda^s_t}}{e^{\lambda^s_t} + e^{\lambda^f_t}}, \ \ \ \ \ w^f_t = \frac{e^{\lambda^f_t}}{e^{\lambda^s_t} + e^{\lambda^f_t}},
\end{align}
where $\big[ \cdot \big]$ stands for the concatenation operator.

\subsection{SlowFast and Modalities Fusion Strategies}
\label{sec:modalities_fusion}
\begin{figure}
\centering
  \begin{subfigure}[b]{0.48\columnwidth}
    \includegraphics[width=\linewidth]{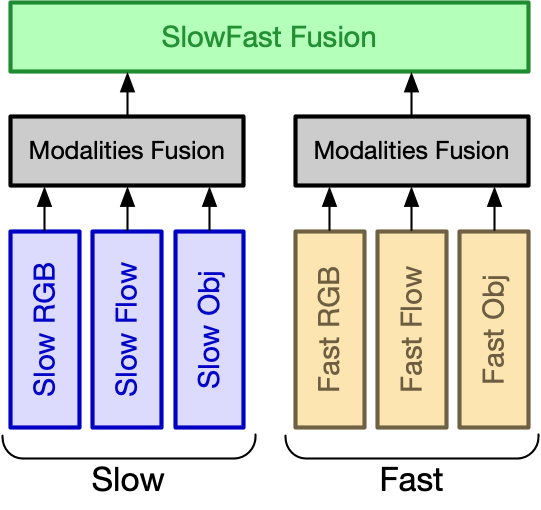}
    \caption{Mod-SF Fusion}
    \label{fig:arc1}
  \end{subfigure}
  \hfill 
  \begin{subfigure}[b]{0.48\columnwidth}
    \includegraphics[width=\linewidth]{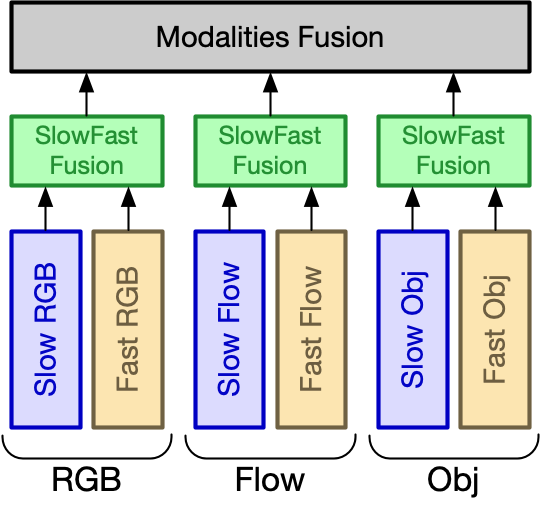}
    \caption{SF-Mod Fusion}
    \label{fig:arc2}
  \end{subfigure}
  \caption{SlowFast and Modalities fusion schemes. (a) Modalities fusion is applied at slow and fast frame rates and then SlowFast fusion is applied to the fused modalities. (b) SlowFast fusion is firstly applied to each modality separately, and then the modalities fusion is applied to the fused time scales.}
    \label{fig:fusion_schemes}
\end{figure}

As proposed in \cite{rulstm}, anticipating future actions can take advantage of multi-modal input representations. 
For this reason, RU-LSTM proposes an attention mechanism (MATT module) that properly weights each input modality. In our work, we exploit the multi-modal video representation and investigate two different techniques to embed both multi-modal and multi-scale inputs. As shown in Figure~\ref{fig:fusion_schemes}, we could either merge our modalities with a MATT module and then fuse both slow and fast branches~(Fig.~\ref{fig:arc1}), or firstly fuse slow and fast branches for each modality, and then merge with a MATT module the multi-modal representations (Fig.~\ref{fig:arc2}). More specifically, Figure~\ref{fig:arc1} depicts an architecture that fuses two RU-LSTMs trained on two different time scales with our slow-fast attention scheme. The input of the attention network is the concatenation of the time scale branches, where each branch is represented by the weighted internal representation $\bm{r}_t$ of the R-LSTM encoders for all the modalities, using the pre-trained modalities attention weights.

As discussed in Sec~\ref{sec:slowfast-rulstm}, in Figure~\ref{fig:arc2} each modality is trained with a slow and fast branch, fused with the slow and fast module and then each modality is merged with the same MATT used in RU-LSTM.

\section{Experimental Results}
\label{sec:results}

We conduct several experiments on two popular datasets used for action anticipation in order to investigate our SlowFast RULSTM model. Furthermore, we study two architectures that embed different fusion mechanisms dealing with multi-modal and multi-scale inputs. In the following, we describe our datasets, metrics and performed experiments in order to show the impact of our slow and fast modelling approach.

\paragraph{Datasets} We experiment on two popular egocentric datasets: EpicKitchens-55~\cite{EPIC-KTICHENS} and EGTEA Gaze+~\cite{egtea}. EpicKitchens-55 collects $55$ hours of recorded videos and $39,596$ annotations of $32$ participants involved in their daily kitchen activities. The annotations contain $125$ verb and $352$ noun classes. All unique (\textit{verb}, \textit{noun}) pairs are considered for a total of $2,513$ unique action labels. EGTEA Gaze+ contains $28$ hours of video clips showing hand-object interaction actions performed by $32$ participants. It contains $19$ verbs, $51$ nouns and $106$ unique actions. The average across three splits reported by the authors of the dataset is considered. 

\paragraph{Evaluation Metric} 
For both datasets, we evaluate our proposed SlowFast RULSTM model using Top-5 accuracy metric at different anticipation times.

\paragraph{Implementation Details} 
We use PyTorch~\cite{NEURIPS2019_9015} for our implementation and use pre-extracted features provided by~\cite{rulstm} for training our method. We found beneficial to train each branch separately and then fine-tuning at the fusion stages. Specifically, for Mod-SF Fusion approach (see~\ref{fig:arc1}), we train RU-LSTM at different frame rates using its standard training pipeline and then fine-tune slow and fast branches at the final stage. For SF-Mod Fusion approach, we apply a similar training strategy.

\subsection{Quantitative Results}

\paragraph{Evaluation Results on EpicKitchens-55}
Table~\ref{tab:rulstm_vs__action_acc_epic-kitchens} reports our results for SlowFast RULSTM and RU-LSTM models on EpicKitchens-55 dataset. Our method outperforms RU-LSTM considering both each modality separately and their fusion. The RGB branch shows an improvement of 1.22\% at $1~s$. Additionally, almost 1\% of improvement is achieved for both FLOW and OBJ modalities. Our model, combining all modalities, achieves a 36.09\% anticipation accuracy at $1~s$, with an improvement of approximately 0.8\% over RU-LSTM baseline. Our  model also shows a remarkable gain at $2~s$ of 1.14\% validating our idea to use a multi-scale approach for capturing more information at the early stages of action anticipation. Our results prove that processing egocentric videos at different frame rates improves the prediction accuracy.

\begin{table}[t]
    \centering
    \begin{adjustbox}{width=\linewidth,center}
    \setlength{\tabcolsep}{3pt}
     \begin{tabular}{l l c c c c}
         \hline
         \multicolumn{6}{c}{Top-5 ACTION Accuracy\% @ different \(\tau_a\)(s)} \\
         \hline
          & & \hspace*{5mm} 2.0 & \hspace*{5mm} 1.5 & \hspace*{5mm} 1.0 & \hspace*{5mm} 0.5 \\
         \hline
         \multirow{2}{*}{RGB} & RULSTM\cite{rulstm} &  \hspace*{5mm}25.44 & \hspace*{5mm}28.32 & \hspace*{5mm}30.83 & \hspace*{5mm}33.31\\ 
         
         & SF-RULSTM &  \hspace*{5mm}\textbf{26.78} & \hspace*{5mm}\textbf{29.25} & \hspace*{5mm}\textbf{32.05} & \hspace*{5mm}\textbf{34.34}\\
         \hline
         & Imp. & \hspace*{5mm}+1.34 & \hspace*{5mm}+0.93 & \hspace*{5mm}+1.22 & \hspace*{5mm}+1.03\\
         \hline
         \hline
         
         \multirow{2}{*}{FLOW} & RULSTM\cite{rulstm} &  \hspace*{5mm}17.38 & \hspace*{5mm}18.91 & \hspace*{5mm}21.42 & \hspace*{5mm}23.49\\ 
         
         & SF-RULSTM &  \hspace*{5mm}\textbf{18.01} & \hspace*{5mm}\textbf{19.82} & \hspace*{5mm}\textbf{22.36} & \hspace*{5mm}\textbf{24.15}\\
         \hline
         & Imp. & \hspace*{5mm}+0.63 & \hspace*{5mm}+0.91 & \hspace*{5mm}+0.94 & \hspace*{5mm}+0.66\\
         \hline
         \hline

         \multirow{2}{*}{OBJ} & RULSTM\cite{rulstm} &  \hspace*{5mm}24.56 & \hspace*{5mm}26.61 & \hspace*{5mm} 29.89 & \hspace*{5mm}31.82\\ 
         
         & SF-RULSTM &  \hspace*{5mm}\textbf{25.61} & \hspace*{5mm}\textbf{27.64} & \hspace*{5mm}\textbf{30.8} & \hspace*{5mm}\textbf{32.15}\\
         \hline
         & Imp. & \hspace*{5mm}+1.05 & \hspace*{5mm}+1.03 & \hspace*{5mm}+0.91 & \hspace*{5mm}+0.33\\
         \hline
         \hline
         
          \multirow{2}{*}{FUSION} & RULSTM\cite{rulstm} &  \hspace*{5mm}29.44 & \hspace*{5mm}32.24 & \hspace*{5mm}35.32 & \hspace*{5mm}37.37\\ 
         
         
         & SF-RULSTM &  \hspace*{5mm}\textbf{30.58} & \hspace*{5mm}\textbf{32.83} & \hspace*{5mm}\textbf{36.09} & \hspace*{5mm}\textbf{37.87}\\
         \hline
         & Imp. & \hspace*{5mm}+1.14 & \hspace*{5mm}+0.59& \hspace*{5mm}+0.77 & \hspace*{5mm}+0.5\\
         \hline
         
    \end{tabular}
    \end{adjustbox}
    \caption{Top-5 accuracy at different anticipation times for RU-LSTM and our SF-RULSTM model.}
    \label{tab:rulstm_vs__action_acc_epic-kitchens}
\end{table}

Table \ref{tab:TAB_comarison} reports a comparison between SlowFast RULSTM and Temporal Aggregation Block (TAB) models,  as proposed in \cite{tab}, which is a current state-of-the-art multi-scale approach for action anticipation. We report results at anticipation accuracy of $1~s$, as authors do not provide anticipation accuracy at different anticipation times. TAB performance is obtained by using the same configuration reported in \cite{tab}. Our results show an accuracy improvement for both RGB and FLOW modalities of +3.8\% and +2.76\%, respectively. In this case, our improvement for both OBJ modality and complete model is less marked, yet our slow-fast fusion model still outperforms TAB model.

\begin{table}[t]
    \centering
    \begin{adjustbox}{width=\linewidth,center}
    \setlength{\tabcolsep}{11pt}
    \begin{tabular}{l|c|c|c|c}
         \hline
         \multicolumn{5}{c}{Top-5 ACTION Accuracy\% @ 1s} \\
         \hline
         &  RGB & FLOW & OBJ & FUSION \\
         \hline
         TAB & 28.25 & 19.60 & 30.09 & 35.73\\
         SF-RULSTM & \textbf{32.05} & \textbf{22.36} & \textbf{30.8} & \textbf{36.09}\\
         \hline
         \hline
         Imp. & +3.8 & +2.76 & +0.71 & +0.36\\
         \hline
    \end{tabular}
    \end{adjustbox}
    \caption{Comparison of action anticipation Top-5 accuracy at $1~s$ between SF-RULSTM  and TAB~\cite{tab} model.}
    \label{tab:TAB_comarison}
\end{table}

\paragraph{Evaluation Results on EGTEA Gaze+}
Table \ref{tab:rulstm_vs__action_acc_egtea} compares our proposed SlowFast RULSTM model to RU-LSTM model on EGTEA Gaze+ dataset. For this dataset, only RGB and optical flow features are available, and we train RU-LSTM model to obtain results for both modalities. By contrast, RU-LSTM fusion results are reported from~\cite{rulstm}. The table shows a maximum improvement for the FLOW modality of approximately +3.5\% at $1~s$. Furthermore, our complete model improves the anticipation accuracy at $1~s$ by 1.2\%, which can be considered a relevant gain due to the reduced number of classes of this dataset compared to EpicKitchens-55 (106 instead of 2513 classes).

\begin{table}[t]
    \centering
    \begin{adjustbox}{width=\linewidth,center}
    \setlength{\tabcolsep}{4pt}
     \begin{tabular}{l l c c c c}
         \hline
         \multicolumn{6}{c}{Top-5 ACTION Accuracy\% @ different \(\tau_a\)(s)} \\
         \hline
          & & \hspace*{5mm} 2.0 & \hspace*{5mm} 1.5 & \hspace*{5mm} 1.0 & \hspace*{5mm} 0.5 \\
         \hline
         \multirow{2}{*}{RGB} & RULSTM &  \hspace*{5mm}56.41 & \hspace*{5mm}60.68 & \hspace*{5mm}66.76 & \hspace*{5mm}72.04\\ 
         
         & SF-RULSTM &  \hspace*{5mm}\textbf{57.84} & \hspace*{5mm}\textbf{62.36} & \hspace*{5mm}\textbf{67.21} & \hspace*{5mm}\textbf{72.32}\\
         \hline
         & Imp. & \hspace*{5mm}+1.43 & \hspace*{5mm}+1.68 & \hspace*{5mm}+0.45 & \hspace*{5mm}+0.28\\
         \hline
         \hline
         
         \multirow{2}{*}{FLOW} & RULSTM &  \hspace*{5mm}33.92 & \hspace*{5mm}35.83 & \hspace*{5mm}39.51 & \hspace*{5mm}42.62\\ 
         
         & SF-RULSTM &  \hspace*{5mm}\textbf{36.93} & \hspace*{5mm}\textbf{39.29} & \hspace*{5mm}\textbf{42.84} & \hspace*{5mm}\textbf{45.94}\\
         \hline
         & Imp. & \hspace*{5mm}+3.01 & \hspace*{5mm}+3.46 & \hspace*{5mm}+3.33 & \hspace*{5mm}+3.32\\
         \hline
         \hline
         
          \multirow{2}{*}{FUSION} & RULSTM\cite{rulstm} &  \hspace*{5mm}56.82 & \hspace*{5mm}\textbf{61.42} & \hspace*{5mm}66.4 & \hspace*{5mm}71.84\\ 
         
         
         & SF-RULSTM &  \hspace*{5mm}\textbf{57.48} & \hspace*{5mm}61.37 & \hspace*{5mm}\textbf{67.6} & \hspace*{5mm}\textbf{72.22}\\
         \hline
         & Imp. & \hspace*{5mm}+0.66 & \hspace*{5mm}-0.05& \hspace*{5mm}+1.2 & \hspace*{5mm}+0.38\\
         \hline
         
    \end{tabular}
    \end{adjustbox}
    \caption{Top-5 accuracy at different anticipation times for EGTEA Gaze+ dataset.}
    \label{tab:rulstm_vs__action_acc_egtea}
\end{table}

         
         
         

         
         
         
         

\subsection{Ablation Experiments on EpicKitchens}
To assess the performance of each part of our model, we conduct a set of ablative experiments. In this case, we focus on EpicKitchens-55 dataset. Additionally, all single modality-related experiments use only RGB features, as they can be assumed to be more inclusive features than both optical flow and object-based features.

\paragraph{Selection of Time Step Value}
The main element of our model is represented by the choice of slow and fast time steps to be fused. Table \ref{tab:alphas_rgb} illustrates our anticipation accuracy using different time steps (\(\alpha \in \{0.1, 0.2, 0.25, 0.5, 1.0\}\)) for RGB features. As shown, the best results (at $1~s$) are obtained selecting \(\alpha = 0.125~s\) and \(\alpha = 0.5~s\). For this reason, we use these two values for our fast and slow branches, respectively.

\begin{figure}[t!]
    \centering
    \includegraphics[width=\columnwidth]{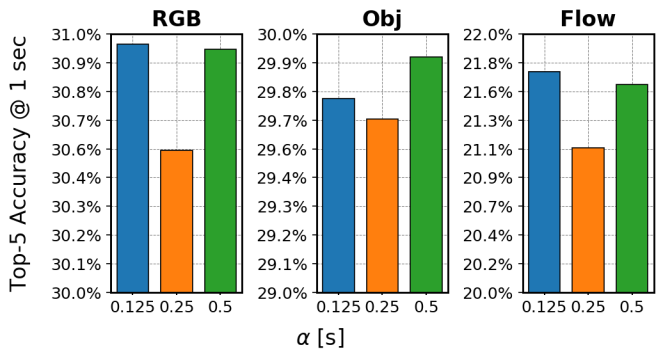}
    \caption{Top-5 accuracy varying the time step $\alpha$ for different input modalities. We select $\alpha \in \{0.125, 0.5\}$ for our SlowFast architecture as each branch appears more accurate with respect to selecting $\alpha=0.25$, as used in~\cite{rulstm}.}
    \label{fig:mod_over_framerate}
\end{figure}

Additionally, Figure \ref{fig:mod_over_framerate} compares Top-5 accuracy results, for each modality, using three different time steps: $0.125$ and $0.5$, as obtained by our previous experiments for the RGB modality, and $0.25$, which represents the default time step value used in \cite{R-2plus1-D}. As shown, our selected time steps improve Top-5 accuracy for each modality.

\begin{table}[t]
    \centering
    \begin{adjustbox}{width=\linewidth,center}
    \setlength{\tabcolsep}{3.5pt}
     \begin{tabular}{c | c c c c c c c c}
         \hline
         & \multicolumn{8}{c}{Top-5 ACTION Accuracy\% @ different \(\tau_a\)(s)} \\
         \hline
         \(\alpha\) (s) & 2.0 & 1.75 & 1.5 & 1.25 & 1.0 & 0.75 & 0.5 & 0.25 \\
         \hline
         0.1 & 25.13 & - & 27.26 & - & 30.44 & - & 33.27 & - \\ 
         
         0.125 & 24.53 & 25.63 & 27.3 & \textbf{28.97} & \textbf{30.96} & \textbf{32.23} & \textbf{33.49} & \textbf{35.02} \\ 
         
         0.2 & 25.16 & - & - & - & 30.71 & - & - & - \\
         
         0.25 & 25.2 & \textbf{25.84} & 27.78 & 28.84 & 30.55 & 31.92 & 33.19 & 34.43 \\
         
         0.5 & \textbf{26.39} & - & \textbf{28.4} & - & \textbf{30.94} & - & 32.87 & - \\
         
         1.0 & 25.56 & - & - & - & 30.13 & - & - & - \\ 
         \hline
    \end{tabular}
    \end{adjustbox}
    \caption{Top-5 accuracy at different time steps (\(\alpha\)) for a single modality (RGB). At $1~s$ the best performance is achieved considering two frame rates: $0.125$ and $0.5$.}
    \label{tab:alphas_rgb}
\end{table}

\paragraph{Sequence Length Encoding}
Extracting relevant features from a video sequence may not only depend on the selected frame rate but also on the length of observed sequences. To this end, we test the impact of different buffer lengths on the anticipation task for the RGB features. Two buffer lengths are considered: \(\tau_e = 1.5~s\), as proposed in \cite{rulstm}, and \(\tau_e = 3.0~s\). As shown in Table \ref{tab:encoding_rgb}, increasing the buffer length provides a noticeable improvement for the slow model (\(\alpha = 0.5~s\)), while the opposite arises for the fast model (\(\alpha = 0.125~s\)). Since the slow model processes a smaller number of video frames, it seems to be able to store more past frames. By contrast, increasing the buffer of the fast model increases its complexity, requiring a smaller window size to achieve better results.

\begin{table}[t]
    \centering
    \begin{adjustbox}{width=\linewidth,center}
    \setlength{\tabcolsep}{28pt}
     \begin{tabular}{c | c c}
         \hline
         & \multicolumn{2}{c}{Top-5 ACTION Accuracy\% @ 1s} \\
         \hline
         \(\tau_e\) (s) & \(\alpha=0.125\) & \(\alpha=0.5\) \\
         \hline
         1.5 & \textbf{30.96} & 30.94\\ 
         
         3.0 &  30.66 & \textbf{31.44}\\
         \hline
    \end{tabular}
    \end{adjustbox}
    \caption{Action anticipation results at $1~s$ for two different lengths of encoding time (\(\tau_e\)) for RGB modality.}
    \label{tab:encoding_rgb}
\end{table}

\paragraph{SlowFast Fusion}
Table \ref{tab:slow_fast_fusion} reports our results for different slow-fast fusion schemes considering the RGB modality. The first three rows shows different fusion methodologies using two scale-branches: slow (with \(\alpha = 0.5s\)) and fast (with \(\alpha = 0.125s\)). We consider two additional fusion techniques other than the proposed attention-based fusion: 

\begin{itemize}
    \item \textit{Concat}: prediction obtained directly from the concatenation of the internal representations of the slow and fast branches;
    \item \textit{Ensemble}: average of the predictions of the slow and fast branches.
\end{itemize} 

As shown, the best fusion scheme at $1~s$ is represented by an attention-based approach, which appears to better discriminate which branch should be used more for predicting future actions. The last row reports our results for the attention-based model considering an additional scale-branch (\(\alpha = 0.25~s\)). These results confirm that anticipating future human actions requires different time scales for obtaining better performance. Among the proposed models, best results are achieved using two scale branches (slow and fast), while adding another branch does not provide any improvement.

\begin{table}[t]
    \centering
    \begin{adjustbox}{width=\linewidth,center}
    \setlength{\tabcolsep}{3pt}
     \begin{tabular}{l c c c c c}
         \hline
         \multicolumn{6}{c}{Top-5 ACTION Accuracy\% @ different \(\tau_a\)(s)} \\
         \hline
         & \(\alpha\)(s) & 2.0 & 1.5 & 1.0 & 0.5 \\
         \hline
          Concat & \{0.125, 0.5\} & 24.59 & 26.9 & 30.04 & 32.73\\ 
         
         Ensemble (AVG) & \{0.125, 0.5\} &  26.98 & 29.59 & 31.71 & 34.2\\
         
         Attention & \{0.125, 0.5\} & 26.78 & 29.25 & \underline{\textbf{32.05}} & 34.34\\
         
         Attention & \{0.125, 0.25, 0.5\} &  26.84 & 29.51 & 31.91 & 33.96\\
         \hline
    \end{tabular}
    \end{adjustbox}
    \caption{Top-5 accuracy at different anticipation times for different slow-fast fusion schemes (RGB modality).}
    \label{tab:slow_fast_fusion}
\end{table}

\paragraph{Modalities Fusion}


\begin{figure*}[h!]
    \centering
    \includegraphics[width=\textwidth]{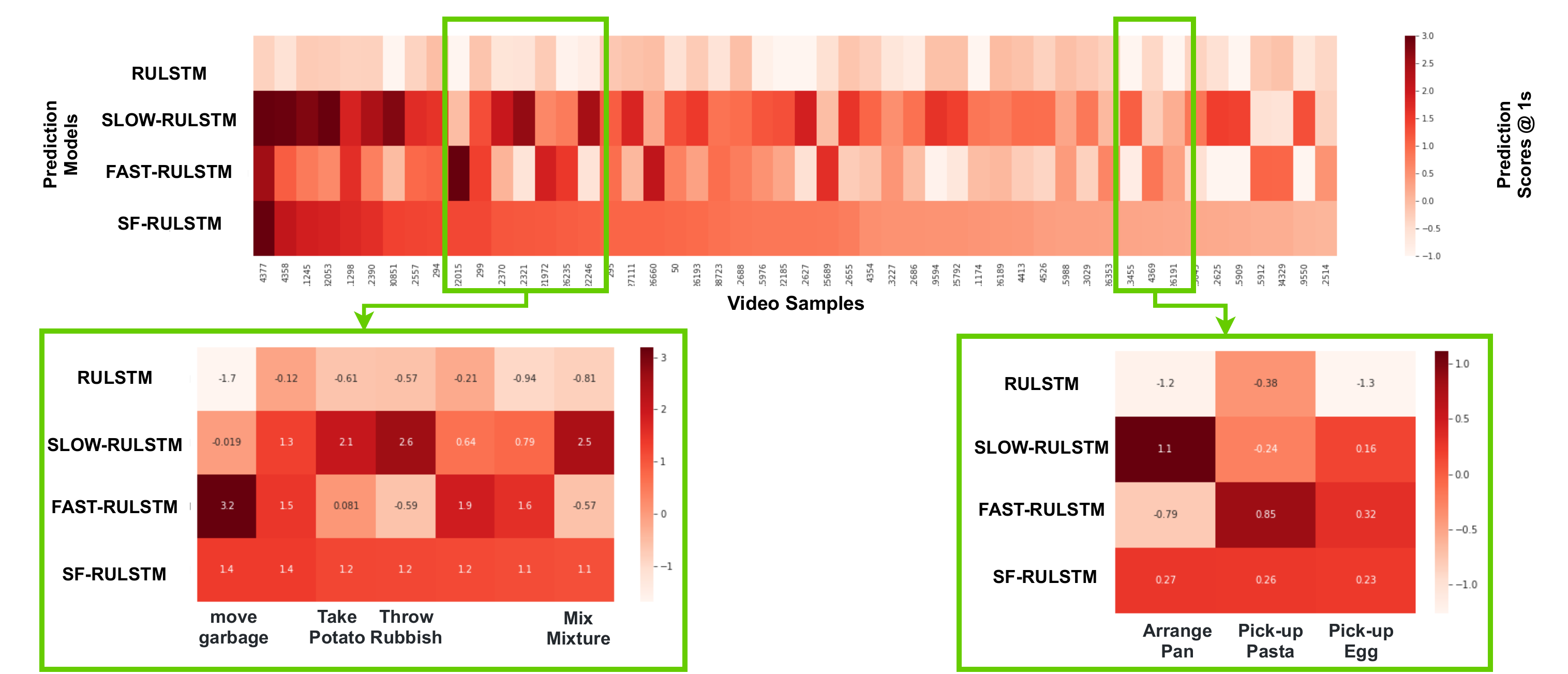}
    \caption{Predictions scores of different video samples from our validation set, where our model provides higher prediction scores than RU-LSTM model. For many actions (\eg, \textit{move garbage, arrange pan)} at least one slow/fast branch has a higher prediction score, and so our complete slow-fast model compared to the selected baseline.}
    \label{fig:samples_scores}
\end{figure*}

\begin{figure*}[h!]
    \centering
    \includegraphics[width=0.95\textwidth]{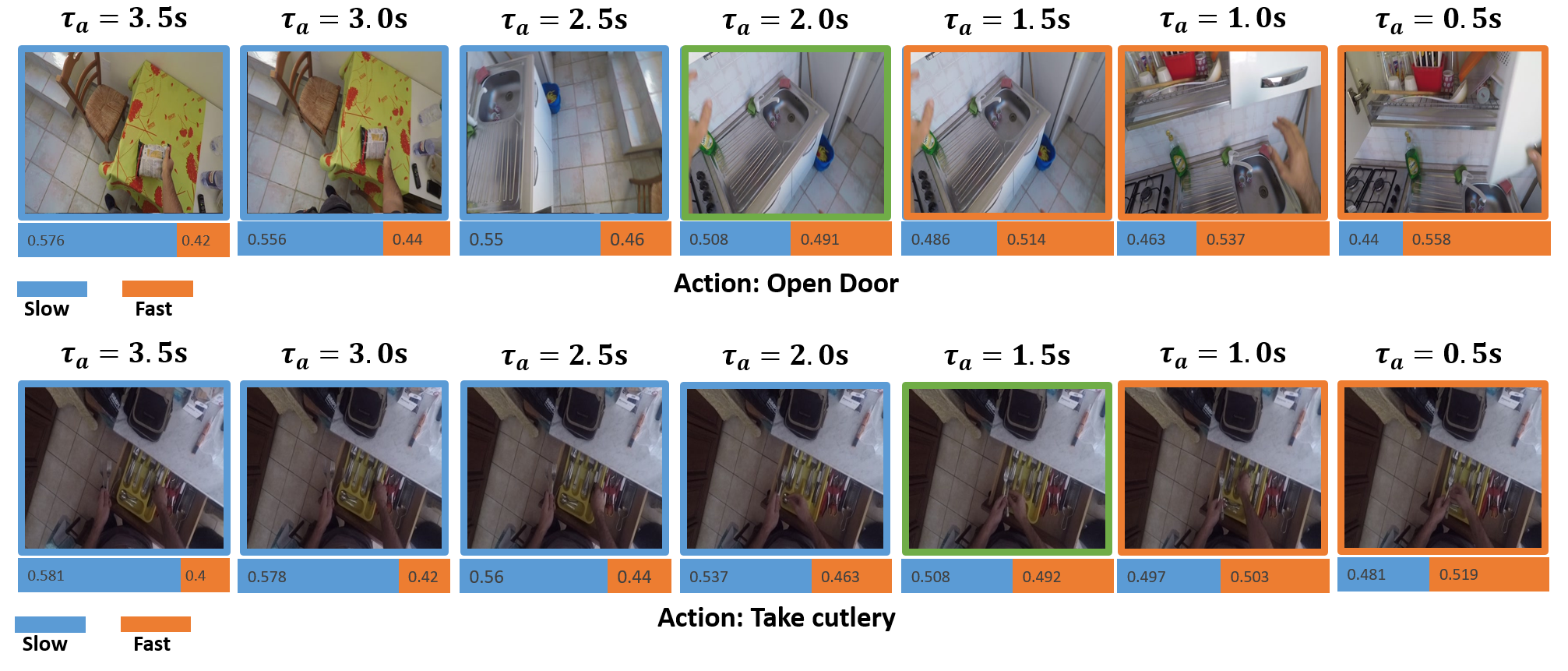}
    \caption{Two examples of actions where slow-fast attention weights change over time. The actions start with no significative changes in the input frames, so the attention mechanism weights more the slow branch. When the action rapidly evolve, more attention is instead provided to the fast branch.}
    \label{fig:attention weights}
\end{figure*}

To assess the performance of the proposed modalities fusion mechanism, shown in Figure \ref{fig:arc1} (Mod-SF Fusion), an alternative fusion architecture (SF-Mod Fusion) is tested (see Figure \ref{fig:arc2}). Table \ref{tab:modalities_fusion} provides Top-5 accuracy for both Mod-SF Fusion and SF-Mod Fusion approaches. Additionally, we change the input of the slow-fast attention to be the concatenation of the internal representations of all R-LSTM branches, instead of the weighting mechanism, as discussed in \ref{sec:modalities_fusion}. As shown, Mod-SF Fusion approach appears the best configuration, since it is easier, compared to the other models, to combine different modalities and allow them to aid each other. Using SF-Mod Fusion, the combination of multi-modal predictions is more complex, and reduces model performance. The approach based on the concatenation, provides the lowest accuracy, which can be due to the huge input size to the attention network.

\begin{table}[h!]
    \centering
    \begin{adjustbox}{width=\linewidth,center}
    \setlength{\tabcolsep}{35pt}
     \begin{tabular}{l|c}
         \hline
         \multicolumn{2}{c}{Top-5 ACTION Accuracy\% @ 1s} \\
         \hline
         Concatenation &  \hspace*{5mm}31.92\\ 
         Mod-SF Fusion &  \hspace*{5mm}\underline{\textbf{36.09}}\\
         SF-Mod Fusion &  \hspace*{5mm}35.28 \\
         \hline
    \end{tabular}
    \end{adjustbox}
    \caption{Top-5 ACTION accuracy at $1~s$ for different variations of modalities fusion.}
    \label{tab:modalities_fusion}
\end{table}

\subsection{Qualitative Results}
We qualitatively evaluate the behaviour of our proposed SF-RULSTM in Figure~\ref{fig:samples_scores} and Figure~\ref{fig:attention weights}. Figure \ref{fig:samples_scores} shows the prediction scores of our SF-RULSTM model (last row) against RU-LSTM model scores (first row) considering a subset of validation samples, \ie, the ones where RU-LSTM assigns low scores. By contrast, our model benefits from either slow (second row) or fast branch (third row) and results in a higher score.

Finally, Figure \ref{fig:attention weights} shows how the slow-fast attention model adapts to different action speeds. Our model is able to select the most appropriate branch for the current action speed, \ie, the slow one, when limited changes in the RGB video stream occur, or the fast branch for actions that evolve more rapidly.

\section{Conclusion}
This work proposes a multi-scale attention-based approach to fuse information extracted at different time scales for anticipating human actions in egocentric videos. Two branches process input videos to capture slow and fast features and better discriminate among different actions (or same action performed by different actors). We design several fusion techniques for combining multiple input modalities and demonstrate that an anticipation model can benefit from fusing input modalities before combining different time scales. 

We outperform a state-of-the-art model on two popular benchmarks, \eg, EpicKitchens-55 and EGTEA GAZE+ and show better results compared to a multi-scale model on EpicKitchens-55 dataset. Our future work will focus on considering more branches and investigating new techniques to better combine several multi-scale branches.
\bigskip

\textit{Acknowledgments:} This research is partially supported by the PRIN-17 PREVUE project, from the Italian MUR (CUP: E94I19000650001).
We gratefully acknowledge the support of NVIDIA for their donation of GPUs, the UniPD DM and CAPRI Consortium for their support and access to computing resources.

{\small
\bibliographystyle{ieee_fullname}
\bibliography{egbib}
}

\end{document}